\documentclass[10pt,twocolumn,letterpaper]{article}

\def\wacvPaperID{arXiv} % Set to 'arXiv' or your Paper ID
\def\confName{WACV}
\def\confYear{2027}

\usepackage[pagenumbers]{wacv} % Forces page numbers for arXiv
\usepackage{amsmath,amssymb}
\usepackage{xcolor}
\usepackage{tikz}
\usetikzlibrary{positioning,arrows.meta,calc,fit,backgrounds}
\usepackage{booktabs}
\usepackage{multirow}
\usepackage{multicol}
\usetikzlibrary{arrows.meta,positioning,fit,backgrounds,calc,shapes.geometric}
\definecolor{cHdr}{RGB}{214,229,247}
\definecolor{cSub}{RGB}{236,242,238}
\definecolor{cBox}{RGB}{248,249,250}
\definecolor{cNorm}{RGB}{250,224,190}
\definecolor{cW}{RGB}{205,222,246}
\definecolor{cAct}{RGB}{193,231,199}
\definecolor{cMM}{RGB}{226,226,230}
\definecolor{cEng}{RGB}{40,90,160}
\definecolor{cData}{RGB}{233,233,233}
\definecolor{cOmg}{RGB}{247,214,214}
\definecolor{cAlloc}{RGB}{206,233,231}
\definecolor{cRef}{RGB}{255,241,178}

\definecolor{wacvblue}{rgb}{0.21,0.49,0.74}
\usepackage[pagebackref,breaklinks,colorlinks,allcolors=wacvblue]{hyperref}

\title{MixFrag: Fragility-Guided Mixed-Precision Post-Training Quantization for Vision Transformers}

\author{
Md. Mehrab Hossain Opi, Robiul Islam Ryad, and Md. Umar Faruk\\
Department of Computer Science and Engineering\\
Khulna University of Engineering \& Technology (KUET)\\
Khulna 9203, Bangladesh\\
{\tt\small opi@cse.kuet.ac.bd, \{ryad2007008, faruk2007016\}@stud.kuet.ac.bd}
}

\begin{document}
\maketitle

\begin{abstract}
Post-training quantization (PTQ) has emerged as an effective solution for deploying Vision Transformers (ViTs) on resource-constrained devices. However, existing PTQ methods typically employ uniform bit-widths across transformer components, overlooking their heterogeneous sensitivity to quantization and leading to inefficient precision allocation. In this paper, we propose \textbf{MixFrag}, a fragility-guided mixed-precision PTQ framework for Vision Transformers. MixFrag first estimates component-level quantization fragility by measuring the Kullback--Leibler (KL) divergence between full-precision and isolated quantized output distributions using a small calibration set. It then formulates bit allocation as a Multiple-Choice Knapsack Problem (MCKP), enabling adaptive layer-wise precision assignment under a target bit budget. Extensive experiments on ImageNet-1K across multiple Vision Transformer architectures demonstrate that MixFrag achieves competitive classification performance under practical mixed-precision settings. Furthermore, evaluations on COCO object detection and instance segmentation show that MixFrag achieves state-of-the-art performance among existing mixed-precision PTQ methods, improving the previous best method by up to \textbf{9.6} AP under the challenging MP3/MP3 setting. Additional analyses validate the proposed fragility metric and demonstrate its strong correlation with the learned bit allocation. These results establish MixFrag as an effective framework for mixed-precision post-training quantization of Vision Transformers.
\end{abstract}
\section{Introduction}
\label{sec:intro}
The remarkable success of the Transformer architecture has revolutionized computer vision, with Vision Transformers (ViTs) achieving state-of-the-art performance across a wide range of tasks, including image classification\cite{liu2021swin, dosovitskiy2021vit}, object detection\cite{zhang2022cat, Dai2021DynamicDE, Zhu2020DeformableDD}, and semantic segmentation\cite{Strudel2021SegmenterTF, Zhou2023OcTrOT}. By modeling long-range dependencies through self-attention, ViTs learn rich and expressive visual representations that often surpass those of convolutional neural networks\cite{He2015DeepRL, Simonyan2014VeryDC}. However, these performance gains come at the cost of high computational complexity, substantial memory requirements, and increased energy consumption. Consequently, deploying ViTs on resource-constrained platforms such as mobile devices, edge systems, and embedded hardware remains a significant challenge\cite{Saha2025VisionTO, Youn2023CompressingVT}.

To facilitate the deployment of Vision Transformers on resource-constrained platforms, numerous model compression techniques have been proposed\cite{Chen2024ComprehensiveSO, Tang2024ASO}, including pruning\cite{Yu2022WidthD, Song2022CPViTCV, Yao2026VPrunerAF}, knowledge distillation~\cite{Habib2023KnowledgeDI, Chen2022DearKDDE, Yang2024ViTKDFK}, low-rank decomposition~\cite{Dong2024LowRankRV, Luo2024SimilarityAwareFL}, and quantization~\cite{Gholami2021ASO, Du2024ModelQA}. Among these techniques, quantization is particularly attractive because it simultaneously reduces model size, memory footprint, and computational cost by representing parameters and activations with low-bit precision. Existing quantization techniques can be broadly categorized into Quantization-Aware Training (QAT)~\cite{Li2022QViTAA, Dong2023PackQViTFS, Choi2018PACTPC} and Post-Training Quantization (PTQ)~\cite{Liu2021PostTrainingQF, nagel2020up, Lv2024PTQ4SAMPQ}. Although QAT generally achieves higher accuracy through end-to-end retraining on the original dataset, it incurs substantial computational cost and often requires access to the complete training data. In contrast, PTQ directly quantizes a pretrained model using only a small unlabeled calibration set, making it significantly more efficient and practical for deploying large-scale pretrained Vision Transformers.

Despite its practicality, PTQ for Vision Transformers remains considerably more challenging than for convolutional neural networks. The self-attention mechanism, LayerNorm, Softmax, and GELU activations exhibit highly asymmetric and layer-dependent distributions, making ViTs particularly sensitive to quantization. Consequently, recent studies have proposed specialized quantization schemes to address these challenges. Representative approaches such as FQ-ViT~\cite{linfq}, RepQ-ViT~\cite{Li2022RepQViTSR}, APQ-ViT~\cite{Ding2022TowardsAP}, and AdaLog~\cite{wu2024adalog} introduce improved quantizers, calibration strategies, and hardware-friendly inference mechanisms to better preserve model accuracy under low-bit quantization. These methods have substantially narrowed the performance gap between quantized and full-precision models, particularly at 4-bit precision. However, maintaining high accuracy under aggressive low-bit settings remains an open challenge, particularly because existing PTQ methods often fail to account for the varying quantization fragility of different transformer components.

Many practical PTQ approaches~\cite{Li2022RepQViTSR, Liu2022PDQuantPQ, weiqdrop} continue to employ uniform bit-width quantization across the network, while existing mixed-precision methods~\cite{Kim2025LampQTA, Ranjan2025LRPQViTMV, Liu2021PostTrainingQF} still struggle to accurately estimate component-specific precision requirements. In practice, transformer components exhibit substantially different levels of fragility under quantization. Allocating identical precision to both fragile and robust components leads to inefficient utilization of the available bit budget. Mixed-precision quantization alleviates this limitation by assigning different bit-widths to different components. However, accurately estimating component fragility and translating it into an optimal global bit allocation remain challenging. Unlike heuristic or proxy-based layer importance measures, fragility directly quantifies the degradation in model outputs caused by quantizing an individual network component, making it a practical and interpretable indicator of precision requirements.

In this paper, we propose MixFrag, a fragility-guided mixed-precision post-training quantization framework for Vision Transformers. Rather than assigning identical precision to all network components, MixFrag explicitly measures the fragility of each quantizable component by estimating its impact on the model's output when quantized in isolation. Based on these fragility scores, we formulate component-wise bit allocation as a Multiple-Choice Knapsack Problem (MCKP), enabling globally optimized precision assignment under a fixed memory budget. The optimized bit allocation is integrated into the AdaLog PTQ framework to produce compact Vision Transformers that achieve higher accuracy than uniform-precision quantization without increasing the average bit-width.

Our main contributions are summarized as follows:

\begin{itemize}
    \item We propose \textbf{MixFrag}, a fragility-guided mixed-precision post-training quantization framework for Vision Transformers that explicitly estimates the quantization fragility of individual network components, enabling precision allocation based on their contribution to model performance rather than uniform bit assignment.

    \item We formulate mixed-precision bit allocation as a \textbf{Multiple-Choice Knapsack Problem (MCKP)}, providing an efficient global optimization strategy that assigns layer-wise bit-widths while satisfying a predefined average bit-width budget.

    \item We perform comprehensive empirical analyses of layer fragility, bit allocation patterns, compression efficiency, and optimization behavior, providing insights into the relationship between quantization fragility and optimal precision assignment.

    \item We extensively evaluate MixFrag on seven Vision Transformer architectures using the ImageNet-1K benchmark. MixFrag consistently improves quantized accuracy over uniform-precision PTQ at identical compression ratios, with the largest gains observed under aggressive 3-bit and 4-bit quantization.
\end{itemize}

The remainder of this paper is organized as follows. Section~\ref{sec:related} reviews related work on post-training quantization and mixed-precision quantization for Vision Transformers. Section~\ref{sec:method} presents the proposed MixFrag framework and the fragility-guided optimization strategy. Section~\ref{sec:experiments} describes the experimental setup and evaluates the proposed method. Finally, Section~\ref{sec:conclusion} concludes the paper and discusses future research directions.

\section{Related Work}
\label{sec:related}

Quantization has emerged as one of the most effective model compression techniques for reducing the memory footprint and computational cost of deep neural networks. Existing quantization approaches can be broadly categorized into Quantization-Aware Training (QAT) and Post-Training Quantization (PTQ). While QAT~\cite{li2023vit, Dong2023PackQViTFS, he2023bivit} generally achieves superior accuracy through end-to-end retraining, it requires access to the original training data as well as substantial computational resources and training time. In contrast, PTQ directly quantizes pretrained models using only a small calibration dataset, making it a considerably more practical and cost-effective solution for real-world deployment. Conventional PTQ methods~\cite{li2021brecq, nagel2020up} have demonstrated remarkable success on convolutional neural networks (CNNs). However, directly extending these techniques to Vision Transformers (ViTs) remains challenging due to the unique characteristics of transformer architectures, including self-attention, LayerNorm, Softmax, and highly irregular activation distributions.

\subsection{Post-Training Quantization for Vision Transformers}

Following the success of PTQ on CNNs, considerable research efforts have been devoted to extending PTQ techniques to transformer-based vision models. Early studies primarily focused on mitigating activation outliers and the quantization sensitivity introduced by self-attention, LayerNorm, Softmax, and GELU. These characteristics make ViTs significantly more sensitive to quantization than convolutional networks, necessitating transformer-specific quantization strategies.

Among the earliest PTQ methods, PTQ4ViT~\cite{yuan2022ptq4vit} introduced Hessian-guided reconstruction to minimize the discrepancy between full-precision and quantized models, demonstrating that second-order information can effectively preserve model accuracy during low-bit quantization. FQ-ViT~\cite{linfq} further addressed the quantization sensitivity of LayerNorm and Softmax by proposing dedicated quantization schemes for these nonlinear operations, substantially improving low-bit performance.

Building upon these ideas, subsequent research focused on refining quantization operators, calibration strategies, and hardware-friendly implementations. Representative methods such as RepQ-ViT~\cite{Li2022RepQViTSR}, APQ-ViT~\cite{Ding2022TowardsAP}, Evol-Q~\cite{frumkin2023jumping}, and AdaLog~\cite{wu2024adalog} progressively improved quantization robustness by introducing more accurate calibration procedures, adaptive quantizers, and efficient optimization strategies. These advances have significantly reduced the accuracy gap between quantized and full-precision Vision Transformers, particularly under 4-bit quantization.

Despite these improvements, existing PTQ methods primarily emphasize the design of better quantizers and calibration mechanisms. Comparatively less attention has been devoted to determining how the limited precision budget should be distributed across different transformer components, motivating the development of mixed-precision quantization methods.
\subsection{Mixed-Precision Quantization}

Uniform-precision quantization assigns an identical bit-width to all network components regardless of their contribution to quantization error. However, Vision Transformers exhibit significant variation in quantization sensitivity across different components, making uniform precision an inefficient use of the available bit budget. Mixed-precision quantization (MPQ) addresses this limitation by assigning different bit-widths to different components, enabling a better trade-off between model accuracy and compression efficiency.

Early MPQ methods were primarily developed for convolutional neural networks, where bit allocation was guided by search-based algorithms, reinforcement learning, Hessian-based sensitivity analysis, and other optimization techniques. Representative methods such as HAQ~\cite{wang2019haq}, HAWQ~\cite{dong2019hawq}, and HAWQ-V2~\cite{dong2020hawq} demonstrated that allocating precision according to component importance consistently outperforms uniform quantization under the same model size constraints.

Building upon these ideas, recent studies have extended mixed-precision quantization to Vision Transformers. Existing ViT MPQ methods estimate component importance using proxy measures, including sensitivity analysis, explainability techniques, and optimization-based formulations. For example, LAMPQ~\cite{Kim2025LampQTA} and Mix-QViT~\cite{ranjan2025mix} formulate mixed-precision allocation as constrained optimization problems, while LRP-QViT~\cite{Ranjan2025LRPQViTMV} employs Layer-wise Relevance Propagation (LRP) to estimate the contribution of individual transformer components for bit allocation. These methods consistently demonstrate that adaptive precision allocation improves quantized accuracy over fixed-bit PTQ while preserving the overall compression ratio.

Despite these advances, accurately estimating the precision requirements of individual transformer components remains an open challenge. Existing approaches primarily rely on indirect proxy measures, such as Hessian approximations, layer importance scores, explainability metrics, or handcrafted sensitivity estimates, which do not explicitly quantify the degradation in model behavior caused by quantization. Consequently, the resulting bit allocations may not faithfully reflect the true quantization fragility of different transformer components.

\subsection{Research Gap}

Although recent PTQ and mixed-precision methods have substantially improved the quantization of Vision Transformers, two key challenges remain. First, existing approaches estimate component importance using indirect proxy measures, such as Hessian approximations, relevance scores, or handcrafted sensitivity metrics, rather than directly measuring the effect of quantization on model behavior. Second, effectively translating these component-wise characteristics into a globally optimized bit allocation under a predefined precision budget remains a challenging optimization problem. Addressing these challenges requires a direct measure of component quantization fragility together with an efficient optimization framework for global precision allocation.
\section{Method}
\label{sec:method}
\subsection{Framework Overview}
\label{sec:overview}

\begin{figure*}[t]
    \centering
    \includegraphics[width=\textwidth]{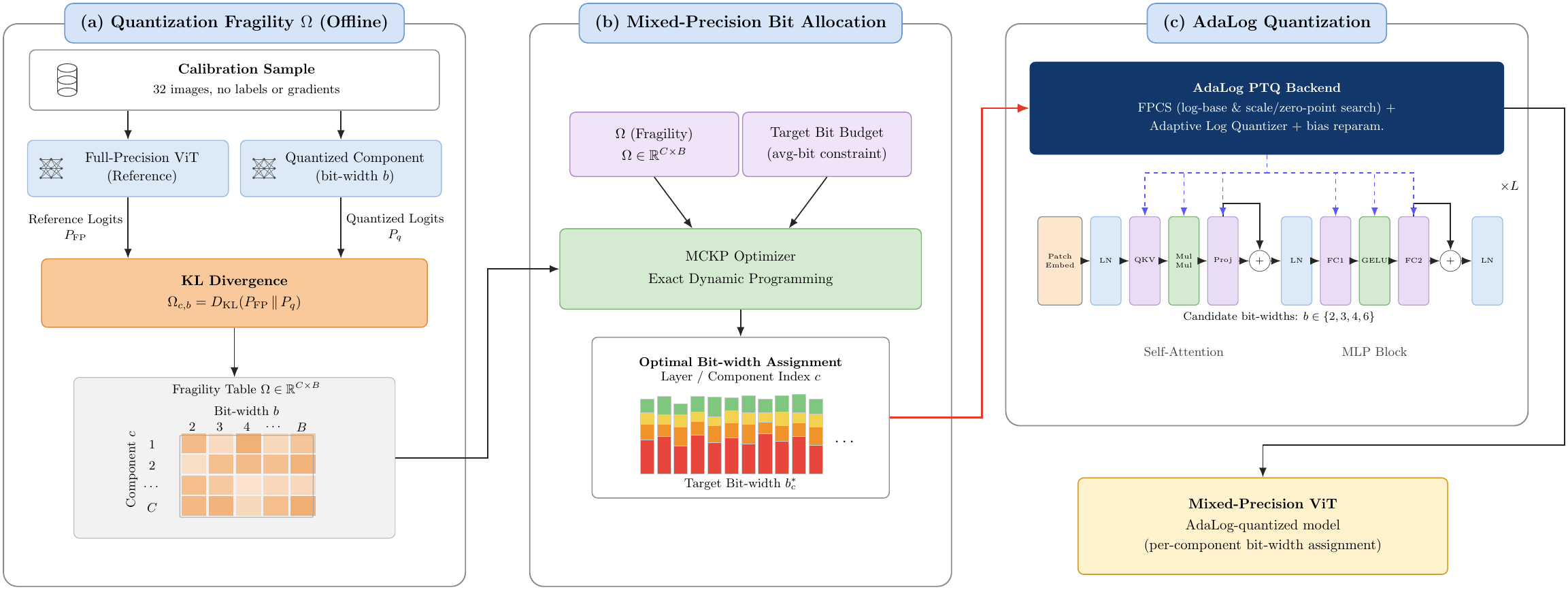}
\caption{\textbf{(a)}~Each quantizable component $c$ is independently quantized at candidate
bit-width $b$ using AdaLog, and its fragility $\Omega_{c,b}$ is computed as the
KL divergence between full-precision and quantized logits on 32 unlabeled
calibration images, forming the fragility table $\Omega\in\mathbb{R}^{C\times B}$.
\textbf{(b)}~The fragility table and target bit budget are optimized via an exact
Multiple-Choice Knapsack Problem (MCKP) solver to obtain the final
per-module allocation $b_c^{*}$.
\textbf{(c)}~AdaLog serves as the quantization backend, where FPCS optimizes
logarithmic bases and scaling factors, while Adaptive Logarithm Quantization
with bias reparameterization handles post-Softmax and post-GELU activations.}
\label{fig:arch}
\end{figure*}

Figure~\ref{fig:arch} presents an overview of the proposed MixFrag framework. Given a pretrained Vision Transformer and a small unlabeled calibration dataset, MixFrag estimates the quantization fragility of individual transformer components, determines an optimal mixed-precision configuration under a predefined bit-width budget, and finally generates a mixed-precision quantized model using the AdaLog post-training quantization backend. The overall framework consists of three stages.
In the first stage, MixFrag estimates the quantization fragility of each quantizable component independently. Specifically, one component is quantized at a candidate bit-width while the remaining components remain in full precision. The output distribution of the partially quantized model is then compared with that of the original full-precision model to measure the performance degradation introduced by quantizing that component. Repeating this process for all components and candidate bit-widths produces a component--bit fragility matrix, which captures the relative sensitivity of different transformer components to quantization.

In the second stage, the computed fragility matrix is used to formulate mixed-precision bit allocation as a Multiple-Choice Knapsack Problem (MCKP). Given a predefined average bit-width budget, the optimization assigns an appropriate precision level to each quantizable component such that the overall quantization fragility is minimized while satisfying the global resource constraint. The resulting solution yields a layer-wise mixed-precision configuration for the entire Vision Transformer.

Finally, the optimized bit allocation is supplied to the AdaLog post-training quantization framework, which performs calibration and quantization according to the assigned precision of each component. Components such as linear projection layers are quantized using their allocated bit-widths, while operations that remain in floating-point precision (e.g., LayerNorm and residual connections) are left unchanged. The output is a mixed-precision Vision Transformer that preserves higher accuracy than uniform-precision quantization under the same average bit-width budget.

\subsection{Quantization Fragility Estimation}

\label{sec:fragility}

The primary challenge of mixed-precision quantization lies in identifying which components require higher numerical precision. Existing approaches\cite{yuan2022ptq4vit, linfq, wu2024adalog} typically estimate component importance using indirect proxy measures, such as Hessian approximations, layer-wise sensitivity analysis, or explainability metrics. While these measures provide useful guidance, they do not explicitly quantify the effect of quantization on the behavior of the model. We instead introduce \textbf{quantization fragility}, which directly measures the degradation in model outputs caused by quantizing an individual component.

Given a pretrained Vision Transformer, let $\mathcal{C}={c_1,c_2,\ldots,c_C}$ denote the set of quantizable components and $\mathcal{B}={b_1,b_2,\ldots,b_B}$ denote the set of candidate bit-widths. In our implementation, quantizable components correspond to the linear projection layers within the self-attention and feed-forward modules, while normalization layers and residual connections remain in floating-point precision.

To estimate the fragility of a component, we adopt a component-wise isolation strategy. For each component $c\in\mathcal{C}$ and candidate bit-width $b\in\mathcal{B}$, only component $c$ is quantized to $b$ bits, while all remaining components are kept in full precision. The partially quantized model and the original full-precision model are then evaluated on the same calibration dataset. This isolation procedure ensures that the measured degradation originates solely from the component under consideration, without interference from other quantized components.

Let $P_{\mathrm{FP}}(\mathbf{x})$ denote the output probability distribution produced by the full-precision model for a calibration sample $\mathbf{x}$, and let $P_{c,b}(\mathbf{x})$ denote the corresponding output distribution when component $c$ is quantized to $b$ bits. The fragility of component $c$ at bit-width $b$ is measured as the average Kullback--Leibler (KL) divergence between these two distributions over the calibration set:
\begin{equation}
  \Omega(c, b) \;=\; \frac{1}{N}\sum_{x\in\mathcal{D}}
  D_{\mathrm{KL}}\!\Bigl(
    \sigma\!\bigl(\mathbf{z}^{\mathrm{fp}}(x)/T\bigr)\,\Big\|\,
    \sigma\!\bigl(\mathbf{z}^{c,b}(x)/T\bigr)
  \Bigr),
  \label{eq:omega}
\end{equation}
where $\mathcal{D}$ is the calibration set ($N$ unlabeled ImageNet
images,
$\mathbf{z}^{\mathrm{fp}}$ and $\mathbf{z}^{c,b}$ are the logits of the
reference and the selectively quantized model, $\sigma$ is the softmax, and $T$ is a temperature. A larger value of $\Omega_{c,b}$ indicates that quantizing component $c$ to bit-width $b$ causes greater deviation from the original model output, implying that the component is more fragile and therefore requires higher precision.

Repeating this procedure for every component and candidate bit-width produces a component--bit fragility matrix $\mathbf{\Omega}\in\mathbb{R}^{C\times B}$, where each entry $\Omega_{c,b}$ stores the fragility score of component $c$ under bit-width $b$. Each row corresponds to a transformer component and each column corresponds to a candidate bit-width. This matrix serves as the input to the mixed-precision optimization stage described in the following subsection.

\subsection{MCKP-Based Mixed-Precision Bit Allocation}

\label{sec:mckp}

The fragility matrix quantifies the sensitivity of each transformer component to different quantization bit-widths. The remaining challenge is to determine an appropriate precision level for every component while satisfying a predefined resource budget. Instead of assigning a uniform bit-width to all components, we formulate this task as a constrained optimization problem that minimizes the overall quantization fragility.

Let $\mathcal{B}={b_1,b_2,\ldots,b_B}$ denote the candidate bit-widths available for each quantizable component. For every component $c_i$, exactly one bit-width must be selected from $\mathcal{B}$. Selecting a higher bit-width generally reduces quantization error but increases the model size, whereas lower bit-widths provide greater compression at the expense of larger accuracy degradation. Therefore, the objective is to achieve the best trade-off between model accuracy and memory efficiency.

We formulate the bit allocation problem as a Multiple-Choice Knapsack Problem (MCKP). In this formulation, each transformer component represents a group of mutually exclusive choices, corresponding to the available bit-widths. The fragility score $\Omega_{i,j}$ is treated as the cost associated with assigning bit-width $b_j$ to component $c_i$, while the selected bit-width contributes to the overall memory budget.

Let $x_{i,j} \in \{0,1\}$ denote a binary decision variable indicating whether bit-width $b_j$ is assigned to component $c_i$. The optimization problem is formulated as

\begin{equation}
\min_{x}
\sum_{i=1}^{C}\sum_{j=1}^{B}
\Omega_{i,j}x_{i,j},
\label{eq:objective}
\end{equation}

subject to

\begin{equation}
\sum_{j=1}^{B}x_{i,j}=1,
\qquad
\forall i,
\label{eq:choice}
\end{equation}

and

\begin{equation}
\sum_{i=1}^{C}\sum_{j=1}^{B}
b_jx_{i,j}
\leq
B_{\mathrm{target}},
\label{eq:budget}
\end{equation}

where $B_{\mathrm{target}}$ denotes the target average bit-width budget. Equation~(\ref{eq:choice}) ensures that each component is assigned exactly one precision level, while Equation~(\ref{eq:budget}) constrains the overall quantization budget.

Since the MCKP is NP-hard, exhaustive search quickly becomes infeasible as the number of transformer components increases. We therefore employ a dynamic programming (DP) algorithm to efficiently obtain the optimal bit allocation under the specified resource constraint. The DP algorithm incrementally considers each transformer component while maintaining the minimum accumulated fragility for every feasible bit budget. This guarantees an optimal layer-wise precision assignment without requiring exhaustive enumeration of all possible combinations.

The optimization produces a mixed-precision configuration specifying the bit-width assigned to every quantizable component. This bit allocation is subsequently passed to the post-training quantization backend, which quantizes each component according to its assigned precision while leaving non-quantized operations unchanged.

\subsection{AdaLog-Based Mixed-Precision Quantization}

\label{sec:adalog}

The optimal bit allocation obtained from the MCKP optimization is integrated into the AdaLog post-training quantization framework to generate the final mixed-precision Vision Transformer. Rather than assigning a uniform precision across the network, each quantizable component is calibrated and quantized according to the bit-width determined by the optimization stage.

Specifically, the quantizable linear projection layers in the self-attention and feed-forward modules are quantized using their assigned precision levels, while operations that are highly sensitive to numerical precision, such as LayerNorm, residual connections, and other non-quantized operations, remain in floating-point precision. During calibration, AdaLog determines the quantization parameters for each component using its adaptive logarithmic quantizer and calibration strategy, thereby minimizing the discrepancy between the quantized and full-precision models.

The proposed framework is independent of the underlying post-training quantization backend. In this work, AdaLog is adopted because of its strong performance on Vision Transformer quantization and its support for flexible layer-wise bit-width assignment. Consequently, MixFrag can be viewed as a backend-agnostic mixed-precision bit allocation framework that can be integrated with other PTQ methods capable of supporting heterogeneous precision configurations.

The resulting model preserves the same average bit-width budget as uniform quantization while allocating precision more effectively according to component fragility, thereby improving the trade-off between model accuracy and compression efficiency.

\section{Experiments}
\label{sec:experiments}

\subsection{Experimental Setup}
\label{sec:setup}

\paragraph{Datasets and Models.}
We evaluate the proposed \textbf{MixFrag} framework on the ImageNet~\cite{Russakovsky2014ImageNetLS} validation set, which comprises 50,000 images spanning 1,000 object categories. Experiments are conducted on three representative Vision Transformer architectures covering diverse model scales and designs: DeiT~\cite{touvron2021training}, ViT~\cite{dosovitskiy2021vit}, and Swin~\cite{liu2021swin}. All pretrained full-precision models are obtained from the \texttt{timm} library\footnote{\url{https://github.com/huggingface/pytorch-image-models}}. To further assess the generalization capability of the proposed bit-allocation strategy beyond image classification, we evaluate MixFrag on the COCO~\cite{Lin2014MicrosoftCC} dataset for object detection and instance segmentation using transformer-based detection frameworks with Swin Transformer backbones, including Mask R-CNN~\cite{he2017} and Cascade Mask R-CNN~\cite{Cai2017CascadeRD}. Unless otherwise specified, all quantized models are evaluated under three quantization configurations: W3/A3, W4/A4, and W6/A6.

\paragraph{Implementation Details.}
Our method is built upon the AdaLog post-training quantization framework, which serves as the underlying quantization backend. Following the standard PTQ protocol, we randomly select 1,024 unlabeled images from the ImageNet-1K training set and 256 unlabeled images from the COCO training set as calibration data for image classification and dense prediction tasks, respectively. For each quantizable transformer component and candidate bit-width, MixFrag estimates quantization fragility by measuring the KL divergence ~\cite{shlens2014notes} between the output distributions of the full-precision model and the corresponding partially quantized model. The resulting component--bit fragility matrix is then used to formulate mixed-precision bit allocation as a Multiple-Choice Knapsack Problem (MCKP), which is solved using dynamic programming under a specified average bit-width budget. The optimized bit allocation is subsequently supplied to the AdaLog backend to generate the final mixed-precision quantized models. Unless otherwise stated, all experiments are conducted using a single NVIDIA RTX 4070 GPU.

\subsection{Comparison with State-of-the-Art}

We compare the proposed \textbf{MixFrag} framework with recent state-of-the-art post-training quantization (PTQ) methods for Vision Transformers, including AdaLog~\cite{wu2024adalog}, AQViT~\cite{wu2024adalog}, I\&S-ViT~\cite{Zhong2026} DopQ-ViT~\cite{Yang2024DopQViTTD}, FIMA-Q~\cite{Wu2025FIMAQPQ}, LAMPQ~\cite{Kim2025LampQTA}, and representative mixed-precision approaches. Experiments are conducted on the ImageNet-1K validation set using seven representative Vision Transformer backbones spanning different architectures and model scales, including ViT, DeiT, and Swin Transformer.

\begin{table*}[t]
\centering
\caption{Comparison with state-of-the-art post-training quantization (PTQ) methods on the ImageNet-1K validation set. All methods are evaluated under identical weight/activation bit-width budgets. ``MP'' indicates whether a method employs mixed-precision quantization. The best and second-best results under each quantization setting are highlighted in \textbf{bold} and \underline{underline}, respectively.}
\label{tab:imagenet_main}
\small
\setlength{\tabcolsep}{4.5pt}
\begin{tabular}{llccccccccc}
\toprule
\textbf{Method} & \textbf{MP} & \textbf{W/A} & \textbf{ViT-S} & \textbf{ViT-B} & \textbf{DeiT-T} & \textbf{DeiT-S} & \textbf{DeiT-B} & \textbf{Swin-S} & \textbf{Swin-B} & \textbf{Avg} \\
\midrule
Full-Precision & -- & 32/32 & 81.39 & 84.54 & 72.21 & 79.85 & 81.80 & 83.23 & 85.27 & 81.18 \\
\midrule
%%%%%%%%%%%%%%%%%%%%%%%%%%%%%%
% W3/A3
%%%%%%%%%%%%%%%%%%%%%%%%%%%%%%
RepQ-ViT\cite{Li2022RepQViTSR} & $\times$ & 3/3 & 0.10 & 0.10 & 0.10 & 3.27 & 7.57 & 1.37 & 1.07 & 1.94 \\
AdaLog~\cite{wu2024adalog} & $\times$ & 3/3 & 12.63 & 29.42 & 25.70 & 22.82 & 55.90 & 58.12 & 61.54 & 38.02 \\
RQViT\cite{Li2022RepQViTSR} & $\times$ & 3/3 & 6.63 & 2.72 & 15.30 & 40.22 & 64.54 & 20.42 & 23.32 & 24.74 \\
AQViT~\cite{wu2024adalog} & $\times$ & 3/3 & 13.08 & 47.55 & 27.54 & 44.56 & 65.30 & 65.08 & 69.11 & 47.46 \\
I\&S-ViT~\cite{Zhong2026} & $\times$ & 3/3 & 45.16 & 63.77 & 41.52 & 55.78 & 73.30 & 74.20 & 69.30 & 60.43 \\
DopQ-ViT~\cite{Yang2024DopQViTTD} & $\times$ & 3/3 & \underline{54.72} & \underline{65.76} & \underline{44.71} & \underline{59.26} & \underline{74.91} & \underline{74.77} & 69.63 & \underline{63.39} \\
FIMA-Q~\cite{Wu2025FIMAQPQ} & $\times$ & 3/3 & \textbf{64.09} & \textbf{77.63} & \textbf{55.55} & \textbf{69.13} & \textbf{76.54} & \textbf{77.26} & \textbf{78.82} & \textbf{71.29} \\
\cmidrule(lr){1-11}
Mix-AdaLog~\cite{ranjan2025mix} & $\checkmark$ & 3/3 & 18.39 & 36.41 & 32.37 & 29.14 & 59.88 & 66.23 & 67.55 & 44.28 \\
LAMPQ~\cite{Kim2025LampQTA} & $\checkmark$ & 3/3 & 23.06 & 48.53 & 37.54 & 45.38 & 61.44 & 70.91 & \underline{75.82} & 51.81 \\
Mix-AQViT~\cite{ranjan2025mix} & $\checkmark$ & 3/3 & 21.44 & 53.36 & 34.83 & 54.08 & 68.32 & 69.66 & 71.18 & 53.27 \\
\textbf{MixFrag (Ours)} & $\checkmark$ & 3/3 & 11.60 & 31.64 & 31.91 & 35.53 & 56.90 & 67.54 & 61.29 & 42.34 \\
\midrule
%%%%%%%%%%%%%%%%%%%%%%%%%%%%%%
% W4/A4
%%%%%%%%%%%%%%%%%%%%%%%%%%%%%%
RepQ-ViT\cite{Li2022RepQViTSR} & $\times$ & 4/4 & 65.05 & 68.48 & 57.43 & 69.03 & 75.61 & 79.45 & 78.32 & 70.48 \\
AdaLog~\cite{wu2024adalog} & $\times$ & 4/4 & 72.75 & 79.68 & 63.52 & 72.06 & 78.03 & 80.77 & 82.47 & 75.61 \\
RQViT \cite{Li2022RepQViTSR}& $\times$ & 4/4 & 68.25 & 73.54 & 59.06 & 70.78 & 77.40 & 80.55 & 80.04 & 72.80 \\
AQViT~\cite{wu2024adalog} & $\times$ & 4/4 & 74.06 & 81.10 & 64.03 & 74.99 & 79.51 & 81.29 & 83.27 & 76.89 \\
I\&S-ViT~\cite{Zhong2026} & $\times$ & 4/4 & 74.87 & 80.07 & 65.21 & 75.81 & 79.97 & 81.17 & 82.60 & 77.10 \\
DopQ-ViT~\cite{Yang2024DopQViTTD} & $\times$ & 4/4 & \underline{75.69} & 80.95 & 65.54 & 75.84 & \underline{80.13} & 81.71 & 83.34 & 77.60 \\
FIMA-Q~\cite{Wu2025FIMAQPQ} & $\times$ & 4/4 & \textbf{76.68} & \textbf{83.04} & \textbf{66.84} & \textbf{76.87} & \textbf{80.33} & \underline{81.82} & 83.60 & \textbf{78.45} \\
\cmidrule(lr){1-11}
Mix-AdaLog~\cite{ranjan2025mix} & $\checkmark$ & 4/4 & 73.57 & 80.95 & 63.87 & 74.73 & 79.15 & 81.42 & 82.87 & 76.65 \\
LAMPQ~\cite{Kim2025LampQTA} & $\checkmark$ & 4/4 & 74.02 & 81.91 & \underline{65.71} & 75.40 & 79.24 & 81.76 & \underline{83.87} & 77.42 \\
Mix-AQViT~\cite{ranjan2025mix} & $\checkmark$ & 4/4 & 75.61 & \underline{82.56} & 65.24 & \underline{76.20} & \textbf{80.33} & \textbf{82.16} & \textbf{84.14} & \underline{78.03} \\
\textbf{MixFrag (Ours)} & $\checkmark$ & 4/4 & 71.92 & 79.76 & 64.88 & 72.94 & 78.28 & 80.28 & 81.44 & 75.64 \\
\midrule
%%%%%%%%%%%%%%%%%%%%%%%%%%%%%%
% W6/A6
%%%%%%%%%%%%%%%%%%%%%%%%%%%%%%
AdaLog~\cite{wu2024adalog} & $\times$ & 6/6 & 80.91 & 84.80 & 71.38 & 79.39 & 81.55 & 83.19 & 85.09 & 80.90 \\
AQViT~\cite{wu2024adalog} & $\times$ & 6/6 & \underline{80.94} & 84.81 & 71.42 & 79.42 & 81.57 & \underline{83.25} & 85.12 & \underline{80.93} \\
I\&S-ViT~\cite{Zhong2026} & $\times$ & 6/6 & 80.43 & 83.82 & 70.85 & 79.15 & 81.68 & 82.89 & 84.94 & 80.54 \\
DopQ-ViT~\cite{Yang2024DopQViTTD} & $\times$ & 6/6 & 80.52 & 84.02 & 71.17 & 79.30 & \underline{81.69} & 82.95 & 84.97 & 80.66 \\
FIMA-Q~\cite{Wu2025FIMAQPQ} & $\times$ & 6/6 & 80.64 & \underline{84.82} & 71.53 & 79.52 & \textbf{81.74} & 83.19 & 85.01 & 80.92 \\
\cmidrule(lr){1-11}
Mix-AdaLog~\cite{ranjan2025mix} & $\checkmark$ & 6/6 & 80.93 & 84.47 & \underline{71.66} & \underline{79.54} & 81.61 & 83.11 & \underline{85.15} & 80.92 \\
Mix-AQViT~\cite{ranjan2025mix} & $\checkmark$ & 6/6 & \textbf{80.99} & \textbf{84.86} & \textbf{71.70} & \textbf{79.62} & 81.63 & \textbf{83.27} & \textbf{85.18} & \textbf{81.04} \\
\textbf{MixFrag (Ours)} & $\checkmark$ & 6/6 & 79.84 & 84.27 & 70.82 & 78.69 & 80.73 & 82.31 & 84.11 & 80.11 \\
\bottomrule
\end{tabular}
\end{table*}
Table~\ref{tab:imagenet_main} reports the top-1 classification accuracy under 3-, 4-, and 6-bit weight/activation quantization. As expected, all methods experience increasing degradation as the quantization precision decreases. At the challenging 3-bit setting, optimization-intensive approaches, particularly FIMA-Q, achieve the highest classification accuracy. In contrast, MixFrag adopts a lightweight fragility-guided bit allocation strategy that prioritizes efficient mixed-precision assignment over costly optimization, leading to a larger accuracy drop under extremely aggressive quantization. Nevertheless, the performance gap narrows considerably at 4-bit quantization and becomes modest at 6-bit, where MixFrag consistently preserves the majority of the full-precision accuracy across all evaluated architectures. These results indicate that the proposed framework is particularly well suited to practical deployment scenarios, where 4--6 bit quantization typically offers a more favorable balance between accuracy and efficiency.
\begin{table*}[t]
\centering
\caption{Comparison with state-of-the-art PTQ methods on the COCO validation set for object detection and instance segmentation.}
\label{tab:coco}
\small
\setlength{\tabcolsep}{4pt}
\begin{tabular}{lllccccccccc}
\toprule
\multirow{3}{*}{\textbf{Method}} & \multirow{3}{*}{\textbf{MP}} & \multirow{3}{*}{\textbf{W/A}} & \multicolumn{4}{c}{\textbf{Mask R-CNN}} & \multicolumn{4}{c}{\textbf{Cascade Mask R-CNN}} & \multirow{3}{*}{\textbf{Avg}} \\
\cmidrule(lr){4-7} \cmidrule(lr){8-11}
&&& \multicolumn{2}{c}{\textbf{Swin-T}} & \multicolumn{2}{c}{\textbf{Swin-S}} & \multicolumn{2}{c}{\textbf{Swin-T}} & \multicolumn{2}{c}{\textbf{Swin-S}} & \\
\cmidrule(lr){4-5} \cmidrule(lr){6-7} \cmidrule(lr){8-9} \cmidrule(lr){10-11}
&&& AP$^{b}$ & AP$^{m}$ & AP$^{b}$ & AP$^{m}$ & AP$^{b}$ & AP$^{m}$ & AP$^{b}$ & AP$^{m}$ & \\
\midrule
Full Precision & -- & 32/32 & 46.0 & 41.6 & 48.5 & 43.3 & 50.4 & 43.7 & 51.9 & 45.0 & 46.3 \\
\midrule
%%%%%%%%%%%%%%%%%%%%%%
% 3-bit
%%%%%%%%%%%%%%%%%%%%%%
RepQ-ViT\cite{Li2022RepQViTSR} & $\times$ & 3/3 & 0.5 & 0.5 & 1.9 & 1.3 & 0.7 & 0.7 & 1.3 & 1.2 & 1.0 \\
AdaLog~\cite{wu2024adalog} & $\times$ & 3/3 & 12.6 & 11.4 & 21.0 & 19.4 & 20.8 & 15.6 & 25.6 & 19.7 & 18.3 \\
AQViT~\cite{wu2024adalog} & $\times$ & 3/3 & 21.1 & 20.3 & 30.7 & 24.6 & 30.9 & 26.2 & 32.3 & 27.4 & 26.7 \\
\cmidrule(lr){1-12}
LRP-RQViT~\cite{Ranjan2025LRPQViTMV} & $\checkmark$ & 3/3 & 5.4 & 3.9 & 11.4 & 10.8 & 7.0 & 7.1 & 13.7 & 13.1 & 9.1 \\
LRP-AQViT~\cite{Ranjan2025LRPQViTMV} & $\checkmark$ & 3/3 & 28.2 & 26.1 & 33.2 & 29.4 & 33.2 & 31.1 & 37.9 & 31.5 & 31.3 \\
\textbf{MixFrag (Ours)} & $\checkmark$ & 3/3 & \textbf{38.8} & \textbf{36.2} & \textbf{42.2} & \textbf{38.9} & \textbf{44.0} & \textbf{38.8} & \textbf{46.8} & \textbf{41.1} & \textbf{40.9} \\
\midrule
%%%%%%%%%%%%%%%%%%%%%%
% 4-bit
%%%%%%%%%%%%%%%%%%%%%%
RepQ-ViT\cite{Li2022RepQViTSR} & $\times$ & 4/4 & 36.1 & 36.0 & 44.2 & 40.2 & 47.0 & 41.4 & 49.3 & 43.1 & 42.2 \\
AdaLog~\cite{wu2024adalog} & $\times$ & 4/4 & 39.1 & 37.7 & 44.3 & 41.2 & 48.2 & 42.3 & 50.6 & 44.0 & 43.4 \\
AQViT~\cite{wu2024adalog} & $\times$ & 4/4 & 41.8 & 39.6 & 45.4 & 41.8 & 48.8 & 42.5 & 50.9 & 44.2 & 44.4 \\
\cmidrule(lr){1-12}
LRP-RQViT~\cite{Ranjan2025LRPQViTMV} & $\checkmark$ & 4/4 & 42.8 & 39.1 & 46.8 & 41.3 & 48.0 & 42.1 & 50.2 & 44.3 & 44.3 \\
LRP-AQViT~\cite{Ranjan2025LRPQViTMV} & $\checkmark$ & 4/4 & 42.9 & 39.9 & 46.8 & 42.2 & 49.3 & 43.0 & 51.1 & 44.4 & 45.0 \\
LAMPQ~\cite{Kim2025LampQTA} & $\checkmark$ & 4/4 & 39.8 & 38.4 & 44.9 & 41.8 & 49.0 & 43.1 & 51.1 & 44.5 & 44.1 \\
\textbf{MixFrag (Ours)} & $\checkmark$ & 4/4 & \textbf{44.1} & \textbf{40.4} & \textbf{47.1} & \textbf{42.7} & 49.2 & \textbf{43.0} & 51.0 & 44.3 & \textbf{45.2} \\
\bottomrule
\end{tabular}
\end{table*}
To further evaluate the effectiveness of MixFrag beyond image classification, we compare against recent PTQ methods on downstream object detection and instance segmentation tasks using the COCO validation set. As shown in Table~\ref{tab:coco}, MixFrag consistently demonstrates superior transfer performance. Under the challenging MP3/MP3 setting, MixFrag achieves an average AP of \textbf{40.9}, substantially outperforming the previous best mixed-precision method (LRP-AQViT, 31.3 AP) by \textbf{9.6} AP points while consistently improving both object detection and instance segmentation across Mask R-CNN and Cascade Mask R-CNN with Swin-T and Swin-S backbones. At MP4/MP4, MixFrag again achieves the best overall performance, obtaining an average AP of \textbf{45.2}, slightly surpassing LRP-AQViT (45.0 AP) and all other competing approaches. Notably, MixFrag is the only method that consistently achieves state-of-the-art performance across both Mask R-CNN and Cascade Mask R-CNN under mixed-precision quantization, demonstrating the robustness of the proposed fragility-guided allocation strategy across different downstream vision tasks.

The contrasting trends between ImageNet classification and COCO downstream tasks suggest that maximizing ImageNet top-1 accuracy alone does not necessarily translate into superior transfer performance after quantization. We hypothesize that the proposed fragility-guided mixed-precision allocation better preserves intermediate feature representations that are critical for dense prediction tasks, resulting in stronger robustness after fine-tuning in detection and segmentation pipelines. Overall, these results demonstrate that MixFrag provides an effective mixed-precision PTQ framework capable of maintaining competitive classification performance while establishing strong downstream transferability across diverse Vision Transformer architectures.

\subsection{Analysis of Layer-wise Bit Allocation}
\label{sec:bitallocation}
% \begin{table}[t]
% \centering
% \caption{Mean assigned bit-width by component type and network stage
% (DeiT-Tiny; block components only for the stage means).}
% \label{tab:typebits}
% \small
% \setlength{\tabcolsep}{3pt}
% \begin{tabular}{@{}lcccccccc@{}}
% \toprule
% & \multicolumn{4}{c}{\textbf{By type}} &
% \multicolumn{3}{c}{\textbf{By stage (blocks)}} \\
% \cmidrule(lr){2-5}\cmidrule(lr){6-8}
% \textbf{Budget} & qkv & proj & fc1 & fc2 & 0--3 & 4--7 & 8--11 \\
% \midrule
% W3 & 3.33 & 3.33 & 2.92 & 2.75 & 3.50 & 3.00 & 2.75 \\
% W4 & 4.17 & 4.33 & 4.17 & 3.50 & 4.38 & 3.94 & 3.81 \\
% W6 & 5.92 & 6.92 & 6.08 & 5.75 & 6.25 & 6.06 & 6.19 \\
% \bottomrule
% \end{tabular}
% \end{table}

\begin{table}[t]
\centering
\caption{Average bit-width assigned by \textbf{MixFrag} to different transformer component types and network stages on DeiT-Tiny under three target bit budgets. Stage-wise averages are computed over the transformer encoder blocks.}
\label{tab:typebits}
\small
\setlength{\tabcolsep}{4pt}
\renewcommand{\arraystretch}{1.15}
\begin{tabular}{@{}lccccccc@{}}
\toprule
\multirow{2}{*}{\textbf{Budget}} &
\multicolumn{4}{c}{\textbf{Component Type}} &
\multicolumn{3}{c}{\textbf{Transformer Stage}} \\
\cmidrule(lr){2-5}\cmidrule(lr){6-8}
& \textbf{QKV} & \textbf{Proj} & \textbf{FC1} & \textbf{FC2}
& \textbf{0--3} & \textbf{4--7} & \textbf{8--11} \\
\midrule
W3/A3 &
\textbf{3.33} &
\textbf{3.33} &
2.92 &
2.75 &
\textbf{3.50} &
3.00 &
2.75 \\

W4/A4 &
4.17 &
\textbf{4.33} &
4.17 &
3.50 &
\textbf{4.38} &
3.94 &
3.81 \\

W6/A6 &
5.92 &
\textbf{6.92} &
6.08 &
5.75 &
\textbf{6.25} &
6.06 &
6.19 \\
\bottomrule
\end{tabular}
\end{table}
Table~\ref{tab:typebits} analyzes the bit allocation produced by the proposed MCKP optimizer by grouping layers according to transformer component type and network stage. Rather than assigning a uniform precision across all layers, MixFrag automatically distributes the available bit budget according to the estimated fragility of individual components, resulting in several consistent allocation patterns across different quantization budgets.

Under the most constrained budget (W3/A3), attention-related operators receive the highest average precision. Specifically, both the QKV and output projection layers are assigned an average of 3.33 bits, whereas the MLP layers (FC1 and FC2) are quantized more aggressively with average precisions of 2.92 and 2.75 bits, respectively. This allocation indicates that preserving higher precision for attention modules is more beneficial than uniformly distributing the available bit budget, even though the MLP layers contain a substantially larger fraction of the model parameters.

A similar trend is observed across transformer stages. The first four transformer blocks receive the highest average precision (3.50 bits), while the middle and final stages are assigned progressively lower bit-widths. This behavior suggests that the fragility metric identifies the early feature extraction layers as being more sensitive to quantization, leading the optimizer to preferentially preserve their numerical precision under tight bit budgets.

As the available precision increases, the allocation naturally becomes more balanced. Under W4/A4, the differences among component types and network stages are noticeably reduced, while under W6/A6 nearly all components receive bit-widths close to the target average. Nevertheless, the projection layer consistently receives the highest precision across all three budgets, indicating that MixFrag repeatedly identifies it as one of the most quantization-sensitive components.

Overall, these observations demonstrate that the proposed fragility-guided optimization does not simply allocate bits according to parameter count or network depth. Instead, it learns a structured allocation strategy that consistently prioritizes the components and stages estimated to have the greatest influence on post-quantization model performance.
\subsection{Quantization Fragility Analysis}
\label{sec:fragility_res}

To validate the effectiveness of the proposed KL-based fragility metric, we analyze both the distribution of component fragility and its relationship with the bit-width allocation produced by MixFrag. Table~\ref{tab:top-fragile} lists the ten most fragile transformer components identified under isolated low-bit quantization, while Table~\ref{tab:corr} reports the statistical correlation between the estimated fragility scores and the final bit-width assignments across different quantization budgets.
\begin{table}[t]

\centering
\caption{Top-10 most fragile transformer components on DeiT-Tiny ranked by the proposed KL-based fragility score under isolated 2-bit quantization. The last three columns report the bit-width assigned by the proposed MCKP optimizer under different average precision budgets.}
\label{tab:top-fragile}
\small
\footnotesize                 % <-- font size
\setlength{\tabcolsep}{2.5pt}   % <-- column spacing
\renewcommand{\arraystretch}{1.15}
\begin{tabular}{@{}c l l c c c c@{}}
\toprule
\textbf{Rank} &
\textbf{Component} &
\textbf{Layer Type} &
$\mathbf{\Omega(c,2)}$ &
\textbf{W3/A3} &
\textbf{W4/A4} &
\textbf{W6/A6} \\
\midrule
1  & Block 0             & FC1          & \textbf{0.504} & 4 & 5 & 7 \\
2  & Patch Emb.          & Projection   & 0.276 & 3 & 4 & 6 \\
3  & Block 0             & FC2          & 0.276 & 4 & 5 & 6 \\
4  & Cls. Head           & Classifier   & 0.170 & 3 & 5 & 6 \\
5  & Block 0             & Proj.        & 0.096 & 4 & 5 & 7 \\
6  & Block 5             & QKV          & 0.093 & 4 & 4 & 6 \\
7  & Block 2             & QKV          & 0.080 & 3 & 5 & 6 \\
8  & Block 3             & QKV          & 0.078 & 4 & 4 & 6 \\
9  & Block 2             & FC1          & 0.075 & 3 & 4 & 6 \\
10 & Block 1             & Proj.        & 0.073 & 4 & 5 & 7 \\
\bottomrule
\end{tabular}
\end{table}
\begin{table}[t]
\centering
\caption{Correlation between the proposed KL-based fragility scores and the bit-width assigned by the MCKP optimizer. Pearson correlation is computed on $\log_{10}\Omega$, while Kendall's coefficient corresponds to the tie-corrected $\tau_b$.}
\label{tab:corr}
\small
\setlength{\tabcolsep}{3pt}
\renewcommand{\arraystretch}{1.0}
\begin{tabular}{@{}lcccc@{}}
\toprule
\textbf{Budget} &
\textbf{Fragility Signal} &
\textbf{Pearson} &
\textbf{Spearman} &
\textbf{Kendall $\tau_b$} \\
\midrule
W3/A3 & $\Omega(c,2)$ & \textbf{0.55} & \textbf{0.57} & \textbf{0.47} \\
W4/A4 & $\Omega(c,3)$ & 0.51 & 0.50 & 0.42 \\
W6/A6 & $\Omega(c,5)$ & 0.07 & 0.02 & 0.01 \\
\bottomrule
\end{tabular}
\end{table}
Table~\ref{tab:top-fragile} reveals several meaningful patterns. The most fragile components are concentrated near the input of the network, including the patch embedding layer and multiple operators in the first transformer block. In particular, the FC1 layer of the first block exhibits the highest fragility score, followed by the patch embedding projection and the corresponding FC2 layer. The classification head also ranks among the most fragile components. These observations indicate that quantization sensitivity is highly non-uniform across the network and cannot be explained solely by layer depth or parameter count, motivating the need for adaptive mixed-precision quantization.

More importantly, the learned bit allocations closely follow the estimated fragility. Highly fragile components consistently receive higher bit-widths across all quantization budgets. For example, the most fragile FC1 layer in the first transformer block is assigned 4, 5, and 7 bits under the W3/A3, W4/A4, and W6/A6 budgets, respectively, while several less fragile components receive substantially fewer bits. This qualitative agreement suggests that the proposed fragility metric provides meaningful guidance for precision allocation.

The statistical analysis in Table~\ref{tab:corr} further supports this observation. Under the most challenging W3/A3 budget, the proposed fragility metric exhibits strong positive correlations with the assigned bit-widths across all three measures, achieving Pearson, Spearman, and Kendall coefficients of 0.55, 0.57, and 0.47, respectively. Similar correlations are maintained under W4/A4, indicating that components estimated to be more fragile are consistently assigned higher precision by the optimizer. As the quantization budget increases to W6/A6, the correlations naturally diminish because most components can already be represented with sufficiently low quantization error, reducing the need for highly selective bit allocation.

Overall, these results validate the proposed fragility metric from both qualitative and quantitative perspectives. The metric identifies intuitive quantization-sensitive components and exhibits a strong correlation with the learned mixed-precision allocation, providing evidence that MixFrag effectively exploits component-level sensitivity when distributing the available precision budget.
\subsection{Ablation Study}
\label{sec:ablation}

To better understand the individual design choices of MixFrag, we conduct ablation studies on the DeiT-Tiny model under different quantization budgets. We first evaluate the effectiveness of the proposed MCKP-based bit allocation strategy and then investigate whether an additional marginal refinement stage provides further performance gains. Unless otherwise stated, all experiments follow the experimental setup described in Section~\ref{sec:setup}, including the same calibration set and quantization configuration.

\subsubsection{Effect of the Allocation Strategy}

To isolate the contribution of the proposed allocation algorithm, we compare the MCKP-based optimizer with two alternative bit allocation strategies while keeping the proposed KL-based fragility metric unchanged. The first baseline randomly assigns candidate bit-widths subject to the target average bit budget, thereby ignoring component sensitivity. The second baseline adopts a greedy strategy that iteratively allocates additional precision to the component with the largest fragility gain per extra bit until the available budget is exhausted.
\begin{table}[t]
\centering
\caption{Comparison of different bit allocation strategies using the proposed KL-based fragility metric.}
\label{tab:allocator_ablation}
\footnotesize
\setlength{\tabcolsep}{5pt}
\renewcommand{\arraystretch}{1.15}
\begin{tabular}{@{}lccc@{}}
\toprule
\textbf{Allocation Strategy} &
\textbf{W3/A3} &
\textbf{W4/A4} &
\textbf{W6/A6} \\
\midrule
Random Allocation & 16.480 & 59.156 & 70.204 \\
Greedy Allocation & 31.908 & 64.826 & 70.606\\
\textbf{MCKP (Ours)} & \textbf{31.91} & \textbf{64.886} & \textbf{70.824} \\
\bottomrule
\end{tabular}
\end{table}
Table~\ref{tab:allocator_ablation} summarizes the results. As expected, random allocation consistently produces the lowest accuracy, demonstrating that simply satisfying the average bit budget is insufficient without considering component fragility. Incorporating the proposed fragility metric through a greedy allocation substantially improves performance across all quantization budgets, confirming the effectiveness of the proposed sensitivity measure.

The proposed MCKP optimizer consistently achieves the best performance. Unlike the greedy strategy, which makes locally optimal decisions at each step, the MCKP formulation jointly optimizes the bit allocation under the global budget constraint and therefore identifies the optimal allocation for the formulated optimization problem. Although the improvement over the greedy baseline is modest, it is consistently observed across all evaluated precision budgets, indicating that the optimization strategy effectively utilizes the available precision budget once informative fragility estimates are available. These results demonstrate that both accurate fragility estimation and globally optimized bit allocation are important for achieving high mixed-precision quantization performance.
\subsubsection{Effect of Marginal In-Context Refinement}

Previous fragility-guided quantization frameworks have employed an additional marginal in-context refinement stage, in which component fragility scores are re-estimated after the initial bit allocation and the optimization problem is solved again using the updated sensitivity estimates. To evaluate whether such iterative refinement benefits the proposed framework, we implement this strategy as an additional optimization step following the initial MCKP allocation.
\begin{table}[t]
\centering
\caption{Effect of the optional marginal in-context refinement stage. All experiments use the proposed KL-based fragility metric and MCKP optimizer.}
\label{tab:refinement}
\footnotesize
\setlength{\tabcolsep}{5pt}
\renewcommand{\arraystretch}{1.15}
\begin{tabular}{@{}lccc@{}}
\toprule
\textbf{Configuration} &
\textbf{W3/A3} &
\textbf{W4/A4} &
\textbf{W6/A6} \\
\midrule
First-pass MCKP (Ours) & 31.908 & 64.886 & 70.824 \\
+ Marginal Refinement  & 29.728 & 64.406 & 70.568 \\
\bottomrule
\end{tabular}
\end{table}
Table~\ref{tab:refinement} compares the proposed single-pass MixFrag framework with its refined counterpart. Across all evaluated quantization budgets, marginal refinement does not improve performance and, in fact, consistently reduces classification accuracy. The degradation is particularly noticeable under the challenging W3/A3 setting, where the refined allocation decreases the top-1 accuracy by more than 2\%. We hypothesize that repeatedly estimating component fragility after quantization introduces additional interactions among components, making the resulting sensitivity estimates less reliable for subsequent bit allocation.

Based on these observations, MixFrag adopts a single-pass fragility estimation followed by one MCKP optimization step. Besides consistently achieving better accuracy, this design also avoids the additional computational overhead associated with iterative refinement, making the overall framework both simpler and more efficient.
\subsection{Discussion}
\label{sec:discussion}

The experimental results highlight two important observations regarding mixed-precision post-training quantization for Vision Transformers. First, quantization sensitivity is highly heterogeneous across transformer components. The fragility analysis shows that only a small subset of layers contributes disproportionately to the overall quantization error, indicating that uniform precision allocation is inherently inefficient. This motivates the use of component-level sensitivity analysis to guide mixed-precision quantization.

Second, the downstream evaluation demonstrates that preserving fragile components is particularly beneficial for dense prediction tasks. Although optimization-intensive PTQ methods achieve higher ImageNet classification accuracy under extremely aggressive quantization, MixFrag consistently delivers superior object detection and instance segmentation performance on the COCO benchmark. These results suggest that preserving critical intermediate representations may be more important for downstream transfer than optimizing classification accuracy alone, emphasizing the importance of evaluating quantization methods beyond ImageNet top-1 accuracy.

Another notable finding is that the proposed fragility metric exhibits a strong correlation with the final bit allocation under constrained quantization budgets. This indicates that the estimated KL divergence provides meaningful guidance for identifying quantization-sensitive components. As the available bit budget increases, the correlation naturally weakens because most components can already be represented with sufficiently small quantization error, reducing the need for highly selective precision allocation. Consequently, the benefits of adaptive mixed-precision quantization become most pronounced under low-bit deployment scenarios.

Despite these encouraging results, several limitations remain. The proposed fragility metric is estimated from a fixed calibration set and evaluates component sensitivity independently, without explicitly modeling interactions among simultaneously quantized components. In addition, the current evaluation focuses on ImageNet classification and representative COCO downstream tasks. Further investigation on larger Vision Transformer backbones, vision foundation models, and additional downstream applications would provide a more comprehensive understanding of the generalization capability of the proposed framework. Finally, although MixFrag is designed as a lightweight single-pass optimization framework, a systematic comparison of runtime, memory consumption, and deployment efficiency with existing optimization-based PTQ methods remains an important direction for future work.
\section{Conclusion}
\label{sec:conclusion}

This paper presented MixFrag, a fragility-guided mixed-precision post-training quantization framework for Vision Transformers. MixFrag estimates component-level quantization fragility using a KL-divergence-based sensitivity metric and formulates bit allocation as a Multiple-Choice Knapsack Problem (MCKP), enabling adaptive precision assignment under a target bit budget through a single-pass optimization framework. Experiments on ImageNet-1K across multiple Vision Transformer architectures demonstrate that MixFrag achieves competitive performance under practical mixed-precision settings. Moreover, evaluations on COCO object detection and instance segmentation show that MixFrag achieves state-of-the-art performance among existing mixed-precision PTQ methods. Additional analyses validate the proposed fragility metric, demonstrate its strong correlation with the learned bit allocation, and confirm the effectiveness of the MCKP optimization strategy. Overall, MixFrag demonstrates that explicitly modeling component-level quantization fragility provides an effective foundation for mixed-precision post-training quantization of Vision Transformers. We hope this work encourages further research on sensitivity-aware quantization methods for efficient deployment of transformer-based vision models.

{
    \small
    \bibliographystyle{ieeenat_fullname}
    \bibliography{main}
}

\end{document}